\documentclass[acmtog]{acmart}
\acmSubmissionID{185}

\usepackage{textgreek}

\usepackage{xcolor}
\usepackage{xspace}
\usepackage{colortbl}

\definecolor{colorTrd}{rgb}{0.95, 0.95, 0.65}
\definecolor{colorSnd}{rgb}{1, 0.85, 0.7}
\definecolor{colorFst}{rgb}{1, 0.7, 0.7}

\definecolor{scenecolor}{rgb}{0.95, 0.95, 0.75}

\usepackage{booktabs} 

\usepackage[ruled]{algorithm2e} 

\SetAlFnt{\small}
\SetAlCapFnt{\small}
\SetAlCapNameFnt{\small}
\SetAlCapHSkip{0pt}

\acmJournal{TOG}

\usepackage{xcolor} 
\usepackage{colortbl}
\usepackage{xspace}
\usepackage{hyperref}
\usepackage[capitalize]{cleveref}
\usepackage{graphicx}
\usepackage{booktabs}
\usepackage{multirow}
\usepackage{multicol}
\usepackage{makecell}
\usepackage{amsmath}
\usepackage{datetime}
\usepackage{scrtime}
\usepackage{pifont}
\usepackage{tikz}
\usepackage{bbm}

\usepackage[shortlabels]{enumitem}

\DeclareMathDelimiter{(}{\mathopen} {operators}{"28}{largesymbols}{"00}
\DeclareMathDelimiter{)}{\mathclose}{operators}{"29}{largesymbols}{"01}

\definecolor{blueevent}{RGB}{58, 89, 209}
\definecolor{redevent}{RGB}{209, 89, 58}
\definecolor{isb}{RGB}{124, 134, 65}
\definecolor{sb}{RGB}{128, 53, 14}

\newcommand{\methodname}{PatternGSL}
\newcommand{\datasetname}{PatternGSLData}

\usetikzlibrary{decorations.pathreplacing}

\usepackage{marvosym}

\AtBeginDocument{%
  }

\begin{document}

\title{PatternGSL: A Structured Specification Language for \\Template-Free and Simulation-Ready 3D Garments}


\author{Zhenyang Li}
\authornote{Both authors contributed equally to this research. This work was completed during their internship.}
\affiliation{%
    \institution{The University of Hong Kong}
    \city{Hong Kong}
    \country{Hong Kong SAR, China}
}

\author{Lutao Jiang}
\authornotemark[1]
\affiliation{%
  \institution{The Hong Kong University of Science and Technology (Guangzhou)}
  \city{Guangzhou}
  \country{Guangzhou, China}
}

\author{Yizhou Zhao}
\affiliation{%
  \institution{Carnegie Mellon University}
  \city{Pittsburgh}
  \country{Pittsburgh, PA, USA}
}

\author{Ying-Cong Chen}
\affiliation{%
  \institution{The Hong Kong University of Science and Technology (Guangzhou)}
  \city{Guangzhou}
  \country{Guangzhou, China}
}

\author{Xin Wang}
\authornotemark[2]
\affiliation{
    \institution{LIGHTSPEED}
    \city{Shenzhen}
    \country{Shenzhen, China}
}

\author{Weikai Chen}
\authornote{Corresponding author.}
\authornote{Project Lead.}
\affiliation{%
  \institution{LIGHTSPEED}
  \city{Los Angeles}
  \country{Los Angeles, CA, USA}
}

\author{Yifan Peng}
\authornotemark[2]
\affiliation{%
    \institution{The University of Hong Kong}
    \city{Hong Kong}
    \country{Hong Kong SAR, China}
}


\begin{abstract}
    Reconstructing realistic, physically plausible garments from a single image remains a fundamental challenge. Template-free methods capture surface geometry but lack explicit sewing structure for simulation; while programmatic systems are simulation-ready but constrained by predefined templates. This reveals a fundamental representation gap between geometric reconstruction and structured garment construction.
    {We present \methodname{}, a structured garment representation in the form of a template-free and learnable specification language that encodes complete sewing patterns, including panel boundaries, parameterized seams, and explicit stitch topology, in a compact and standardized form. \methodname{} preserves the physical rigor of pattern-based models while removing template dependence, elevating sewing structure as a first-class target for generative modeling.
    We further propose a vision-language framework that predicts \methodname{} specifications directly from a single image and decodes them into garments using lightweight deterministic validity handling, without optimization-based refinement or manual cleanup.}
    In addition, we introduce \datasetname{}, the first large-scale image-to-GSL paired dataset comprising 300K samples with complete sewing pattern annotations, enabling supervised VLM training for structured garment reconstruction.
    {Experiments demonstrate improved pattern accuracy over prior baselines, explicit sewing-structure recovery, reliable cloth simulation, and pattern-level editing through the same deterministic decoding pipeline. Code and data-processing scripts will be released at \url{https://lagrangeli.github.io/PatternGSL/}.}
\end{abstract}



\begin{teaserfigure}
    \centering
    \includegraphics[width=\textwidth]{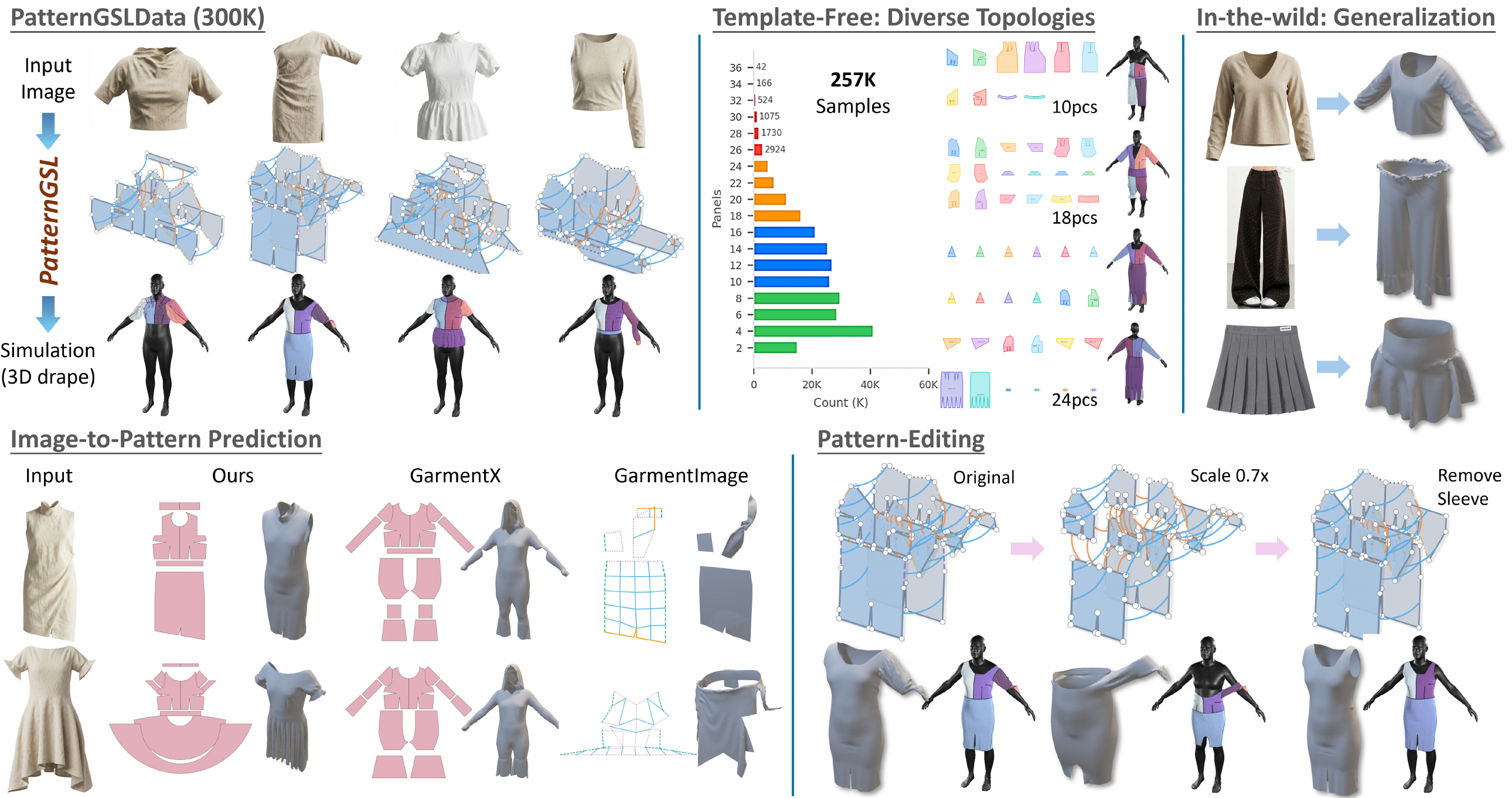}
    \vspace{-0.7cm}
    \caption{{We present \methodname{}, a template-free approach for reconstructing garment sewing patterns from images. Built upon \datasetname{}---a large-scale image-to-pattern dataset with 300K samples---our method learns to predict structured pattern representations that can be directly simulated as 3D garments. In our current dataset and evaluation, \methodname{} covers diverse topologies ranging from 2 to 37 panels and generalizes to in-the-wild photographs.} Compared to prior methods, our approach produces higher-fidelity reconstructions while maintaining simulation compatibility. The predicted patterns further support intuitive editing operations such as scaling and component removal.}
    \label{fig:teaser}
    \vspace{-0.1cm}
\end{teaserfigure}


\maketitle

\section{Introduction}
\label{sec:intro}

Reconstructing realistic and physically plausible garments from images remains a key challenge in digital fashion and 3D reconstruction.
Beyond visual fidelity, practical solutions must seamlessly support physics-based simulation and downstream design workflows. 
Yet many prevailing learning-based approaches~\cite{zhu2020deepfashion3d,saito2019pifu,saito2020pifuhd,corona2021smplicit} represent garments as neural implicit shapes, optimized primarily for visual plausibility.
While such methods can capture diverse shapes without predefined templates, their geometry-first formulations obscure sewing structures and tend to smooth out sharp seams and boundaries, which in turn hinders their direct use in cloth simulation (Fig.~\ref{fig:teaser}).

At the other end of the spectrum, recent programmatic systems for garment modeling highlight the advantages of explicit structure.
GarmentCode~\cite{korosteleva2023garmentcode}, for example, introduces a modular, simulation-ready framework in which garments are constructed through compositional programs that define components and stitching operations. 
This design enables precise control over garment construction and physical behavior. 
Nonetheless, GarmentCode’s expressiveness is rooted in hand-crafted garment programs and predefined components, which constrain the space of admissible topologies. 
Consequently, the framework remains fundamentally template-driven and has limited ability to infer novel sewing structures or arbitrary garment topologies directly from images.


These observations reveal a persistent representation gap. 
Real garments are constructed from 2D fabric panels whose shapes, seam curves, and stitching relationships jointly determine their 3D form and physical behavior.
Bridging this gap calls for a representation that combines the \textbf{physical rigor of pattern-based models} with the \textbf{flexibility of template-free approaches}, while remaining compatible with {learning-based systems}.
Such a representation should explicitly encode sewing structure -- panels, curves, and stitches -- so that reconstructed garments are immediately usable in cloth simulation, yet avoid rigid, hand-crafted templates by supporting arbitrary topologies.
{At the same time, it must be sufficiently regular and standardized to act as a stable, learnable target for image-conditioned generative models.}


To this end, we introduce \textbf{\methodname{}}, a new representation that rethinks how sewing patterns are encoded for learning and simulation.
Rather than treating garments as fixed, hand-crafted programs as in prior work, \methodname{} reframes sewing patterns as a regularized, extensible, and learnable specification.
It distills the essential elements of garment construction into a standardized language in which panels, seam curves, and stitch correspondences are first-class primitives, decoupled from rigid program templates.
In doing so, we preserve the geometric precision and physical validity that make pattern-based approaches appealing, while replacing monolithic programs with a compositional representation that scales to arbitrary garment topologies and can be predicted directly by modern generative models.

Representation-wise, as shown in Fig.~\ref{fig:pipeline}, we organize garments into a clear hierarchy of panels, parameterized edges, and stitches, explicitly separating continuous geometry from discrete topology while preserving their correspondence.
Panel boundaries are encoded with curve parameterizations that regularize shape yet retain high fidelity along seams.
Stitch relationships are structured as explicit pairwise {correspondences}, rather than being inferred heuristically.
Collectively, this design elevates sewing patterns from ad hoc, program-specific artifacts to a consistent, learnable format, providing a stable interface between perception and simulation.
Algorithm-wise, we formulate single-image garment reconstruction as the problem of generating a \methodname{} configuration. 
We train a vision-language model to predict this structured representation directly from an input image, leveraging its ability to generate long, hierarchical outputs that adhere to the constraints of the language.
{The resulting specification is parsed and decoded into a complete sewing pattern with deterministic validity handling, and can then be fed to a cloth simulator without optimization-based refinement or manual cleanup.}

Beyond reconstruction, \methodname{}'s \emph{explicit parameterization} enables direct and intuitive pattern editing. Designers can resize panels by modifying vertex coordinates, reshape boundaries by adjusting curve parameters, or filter panels by removing parts. {The edited specification is then processed by the same deterministic decoder used for reconstruction, preserving compatibility with the simulation pipeline.}
This \emph{editability} naturally extends to in-the-wild scenarios, where patterns predicted from real-world images can be iteratively refined to meet design requirements.
{As shown in Fig.~\ref{fig:teaser}, experiments on synthetic and in-the-wild datasets show that \methodname{} improves pattern accuracy and simulation reliability over prior baselines, achieving 99.2\% simulation success while supporting flexible pattern editing.}
In summary, our contributions are:
\begin{itemize}[leftmargin = 15pt]
    \item A structured and template-free garment specification language, dubbed \methodname{}, that regularizes sewing patterns as a learnable representation, explicitly encoding panels, edges, and stitches while supporting arbitrary garment topologies;
    \item {A vision-language-based generation framework that predicts \methodname{} directly from a single image, together with a parser and deterministic decoder that converts the prediction into complete sewing patterns for simulation;}
    \item \datasetname{}, the first large-scale image-to-GSL paired dataset comprising 300K samples (250K synthetic + 50K photorealistic) with complete panel, edge, and stitch annotations, enabling supervised VLM training;
    \item {Empirical validation on pattern accuracy, stitch correctness, draping reliability, editing validity, and in-the-wild generalization.}
\end{itemize}
\section{Related Work}
\label{sec:related}

\paragraph{3D Garment Reconstruction.}
Template-based approaches~\cite{guan2012drape,pons2017clothcap,santesteban2019learning,vidaurre2020fully} deform predefined meshes using parametric body models~\cite{loper2015smpl,pavlakos2019expressive,xu2020ghum}. Implicit representations~\cite{saito2019pifu,saito2020pifuhd,corona2021smplicit,he2021arch++,xiu2022icon} enable high-resolution digitization but produce fused geometry. BCNet~\cite{jiang2020bcnet} separates layers via depth prediction; DeepFashion3D~\cite{zhu2020deepfashion3d} and MGN~\cite{bhatnagar2019mgn} provide benchmarks. Text-driven methods~\cite{liu2024clothedreamer,srivastava2024wordrobe,huang2024tech,luo2024garverselod} generate garments from descriptions. These output meshes or implicit fields rather than simulation-ready sewing patterns.

\paragraph{Sewing Pattern Representations.}
{Existing approaches show the value of explicit sewing structure but often tie prediction to predefined garment families. NeuralTailor~\cite{korosteleva2022neuraltailor} predicts from point clouds but requires template selection; SewFormer~\cite{liu2023sewformer} uses transformers for panel prediction; GarmentCode~\cite{korosteleva2023garmentcode} provides a powerful programmatic construction system with reusable components and stitching operations. Our work builds on this sewing-based view, but targets a canonicalized sequence representation for learning: panels are deterministically ordered, vertices are preserved as indexed arrays, and stitches explicitly reference panel-edge pairs. Thus PatternGSL is not only a JSON serialization of sewing patterns, but a stable output space for image-conditioned generation.}

{Recent methods also differ in output space and evaluation target. GarmentX~\cite{guo2025garmentx} uses autoregressive garment generation, while our representation keeps explicit panel, edge, and stitch topology as the supervised prediction target. Dress-1-to-3~\cite{li2025dress1to3} recovers 3D garments using diffusion priors and differentiable physics; in contrast, we directly predict 2D sewing patterns as the primary output, enabling topology-aware supervision, pattern editing, and direct cloth simulation.} GarmentImage~\cite{li2024garmentimage} encodes patterns for CNN prediction but loses boundary information; GarmentDiffusion~\cite{li2025garmentdiffusion} generates patterns via diffusion transformers. Physics-based cloth simulation~\cite{narain2012adaptive,tang2018cloth,li2020p} and differentiable simulators~\cite{liang2019differentiable,li2022diffcloth} provide the simulation backends used by many of these systems.

\paragraph{Generative Models and Structured Prediction.}
Diffusion models~\cite{ho2020denoising,song2020denoising,song2020score} achieve remarkable success in image synthesis~\cite{rombach2022high,saharia2022photorealistic} with classifier-free guidance~\cite{ho2022classifier}. For 3D generation, methods span text-to-3D~\cite{poole2022dreamfusion,lin2023magic3d}, multi-view synthesis~\cite{liu2023zero,shi2023mvdream,tang2024lgm}, and meshes~\cite{gao2022get3d,siddiqui2024meshgpt}. Structured outputs include layouts~\cite{inoue2023layoutdm,chai2023layoutdm} and physics-aware motion~\cite{tevet2022human,yuan2023physdiff}. Large VLMs~\cite{bai2023qwen,li2023blip2,alayrac2022flamingo,liu2024llava} combine visual encoders~\cite{dosovitskiy2021vit,radford2021clip} with language models~\cite{touvron2023llama,chiang2023vicuna}, excelling at structured generation~\cite{chen2021pix2seq,raffel2020exploring} with efficient LoRA fine-tuning~\cite{hu2021lora}. We leverage VLMs by designing GSL as a compact structured format.

\section{\methodname{}}
\label{sec:method}

\begin{figure*}[ht]
    \centering
    \includegraphics[width=\linewidth]{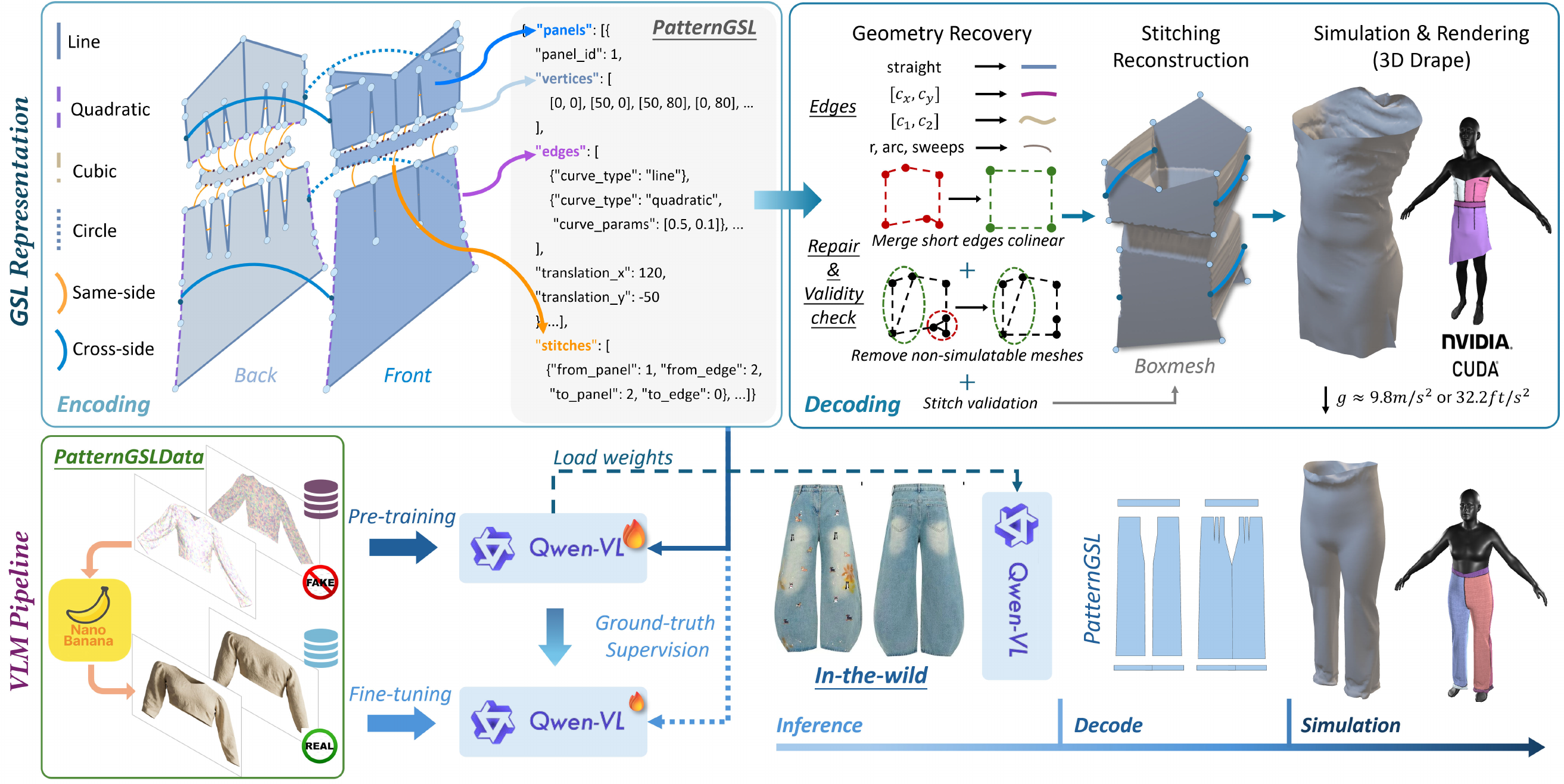}
    \caption{The \methodname{} framework. \textbf{(Top-left)} GSL Representation: legend denotes edge types (line/quadratic/cubic/circle) and stitch types (same-side/cross-side); sewing patterns are encoded into PatternGSL while preserving vertices, curves, and topology. \textbf{(Top-right)} Decoding: geometry recovery, repair (merging short edges, removing invalid panels), stitching reconstruction, boxmesh generation, and physics simulation (NVIDIA Warp). \textbf{(Bottom-left)} VLM Pipeline: \datasetname{} is combined with NanoBanana for back-view synthesis and photorealistic rendering; two-stage training (pre-training on FAKE, fine-tuning on REAL) with ground-truth supervision; fire icons indicate trainable models. \textbf{(Bottom-right)} Inference: frozen Qwen-VL predicts PatternGSL from in-the-wild images, then decoded and simulated into 3D garments.}
    \label{fig:pipeline}
\end{figure*}

\subsection{Design Goal}
\label{sec:design_goal}

{The goal of \methodname{} is to make sewing structure a direct prediction target. A garment pattern is naturally discrete-continuous: panels define 2D fabric pieces, boundary curves define their shape, 3D transforms initialize draping, and stitches define how edges are assembled. We therefore seek a representation that is compact enough for sequence generation, expressive enough for arbitrary panel layouts, and explicit enough to be decoded into a cloth simulation input.}

{PatternGSL follows this organization rather than hiding it in an implicit embedding. Geometry and topology are separated, but their correspondence is kept through indexed references: panels contain ordered vertices and edges, and stitches point to specific panel-edge pairs. This makes the output interpretable, editable, and learnable as a structured language.}

\subsection{Representation Design}
\label{sec:representation}

{We instantiate \methodname{} as a hierarchical JSON structure with three components (Fig.~\ref{fig:pipeline}, top-left): (1) a \texttt{meta} block storing the scale factor $\sigma_{\text{scale}}$, panel count, and boundary sample count $N_s$; (2) a \texttt{panels} array storing geometry and placement; and (3) a \texttt{stitches} array storing topology. The scale factor is part of the representation rather than the output of a separate monocular scale estimator. During training, coordinates are normalized using a fixed scale prior and the corresponding value is stored in the \texttt{meta} fields; at inference, the VLM predicts this scale token jointly with panel geometry and stitch topology.}

\paragraph{Panel Geometry.}
Each panel $\mathcal{P}_p$ is encoded as a JSON object containing {its 2D boundary and 3D placement}. The boundary is represented by an ordered vertex sequence $\{(x_j, y_j)\}_{j=1}^{n_p}$ defining the panel polygon. {To make numeric tokens stable across garment sizes, local coordinates are normalized as}
\begin{align}
    {\mathbf{v}_{\text{gsl}} = \sigma_{\text{scale}} \cdot \mathbf{v}_{\text{local}} .}
\end{align}

For 3D placement initialization during draping simulation, each panel stores six transform parameters: translations $(t_x, t_y, t_z)$ and rotations $(r_x, r_y, r_z)$. The horizontal translations are normalized relative to a reference template width as $(t_x^{\text{gsl}}, t_y^{\text{gsl}}) = (t_x, t_y) \cdot \sigma_{\text{scale}} / W_{\text{ref}}$, while depth and rotations are preserved directly. Each panel carries a binary \texttt{side} attribute (``front'' or ``back'') determined by $\text{sign}(t_z)$, aiding VLM reasoning about garment structure.

\paragraph{Edge and Curve Representation.}
Edges connect consecutive vertices and may be straight or curved. GSL supports four curve types with explicit parameterization. {To avoid absolute control-point coordinates that vary with garment scale, curve parameters are expressed relative to each edge endpoint.} Line segments require no additional parameters. Quadratic B\'{e}zier curves store $(c_0, c_1)$ and reconstruct the control point as $\mathbf{p}_c = \mathbf{p}_0 + (c_0, c_1) \cdot L$, where $L$ is the edge length. {Cubic B\'{e}zier curves store four relative parameters for two control points, and circular arcs store radius and orientation flags.}

While parametric representations suffice for simulation, VLMs benefit from explicit geometric supervision on curve shapes. For curved edges, GSL additionally stores $N_s$ uniformly sampled points along the curve. Critically, straight edges omit sample points entirely, reducing token count substantially since most garment edges are straight, while key curved edges (e.g., necklines, armholes, hemlines) still receive rich geometric supervision.

\paragraph{Stitch Topology.}
Stitch relationships encode how panel edges are sewn together during assembly. Each stitch is represented as a JSON object with a unique \texttt{stitch\_id}, along with source identifiers (\texttt{from\_panel\_id}, \texttt{from\_edge\_index}) and target identifiers (\texttt{to\_panel\_id}, \texttt{to\_edge\_index}) that specify exactly which edges are connected. This explicit encoding avoids ambiguity when geometrically close edges (e.g., front and back necklines) belong to different panels. Unlike implicit representations where stitches must be inferred from geometric proximity, GSL stores exact connectivity, enabling precise topology recovery.

\subsection{{Canonicalized Encoding}}
\label{sec:encoding}

{Before tokenization, we canonicalize each garment specification so that equivalent inputs produce a stable training target. Panel names are sorted to assign deterministic panel IDs from 1 to $P$; vertex ordering within each panel is preserved through indexed arrays; and stitches are rewritten as explicit $(\texttt{panel\_id}, \texttt{edge\_index})$ references. This stable ordering reduces avoidable output ambiguity for the VLM.}

{Given panels $\{(\mathcal{P}_p, \mathbf{T}_p)\}_{p=1}^{P}$ and stitches $\{\mathcal{S}_s\}_{s=1}^{S}$, encoding then applies the normalization above and rounds numeric fields to two decimal places. Edge types and relative curve parameters are inferred from the input curvature fields. For curved edges, $N_s$ boundary samples are evaluated along the parametric curve; for straight edges, samples are omitted. Finally, stitches are converted through the canonical panel-ID mapping and invalid references are discarded.}

\subsection{{Deterministic Decoding and Validity Handling}}
\label{sec:decoding}

{The decoder inverts the structured encoding and performs lightweight deterministic validity handling before simulation (Fig.~\ref{fig:pipeline}, top-right). It first reads the predicted $\sigma_{\text{scale}}$ token from the \texttt{meta} block and applies inverse normalization,}
\begin{align}
    {\mathbf{v}_{\text{local}} = \mathbf{v}_{\text{gsl}} / \sigma_{\text{scale}},}
\end{align}
{with translations recovered analogously. No standalone body-scale estimator is used. The depth $t_z$ is read directly when present or initialized from the predicted \texttt{side} attribute.}

{Edges are reconstructed from the vertex arrays and curve parameters. When predicted curve parameters are missing or corrupted, boundary samples provide a deterministic fallback; for example, a quadratic B\'{e}zier control point can be recovered from the midpoint sample in closed form. Stitches are reconstructed by mapping panel IDs back to panel names. Before simulation, the decoder applies deterministic checks such as merging short collinear edges, removing invalid panels, and validating stitch references. These steps are rule-based sanitation rather than optimization-based refinement or manual correction.}

\subsection{{Representation Properties}}
\label{sec:properties}

{The design gives PatternGSL four properties needed for image-to-pattern learning.}

{\emph{Canonical topology}: sorted panel IDs, preserved vertex order, and explicit stitch references produce a consistent GSL target for a given garment specification. Connectivity is stored directly as index pairs rather than inferred from geometric proximity.}

{\emph{Geometric precision}: coordinate quantization to two decimal places bounds error to $\epsilon < 0.01 / \sigma_{\text{scale}} \approx 0.04$~mm, below typical simulation tolerances ($\sim$0.1~mm). Relative curve parameterization keeps this precision independent of absolute garment scale.}

{\emph{Token efficiency}: storing only boundary vertices yields $O(n)$ tokens per panel, relative curve parameters require 2--4 values per curved edge, and straight edges omit samples. Typical garments encode within $\sim$10K tokens.}

{\emph{Editability}: geometric parameters have direct semantic meaning, so users can modify vertex coordinates, curve parameters, or placements and then run the same deterministic decoder. The complete GSL schema is provided in the supplementary material.}

\begin{figure*}[ht]
    \centering
    \includegraphics[width=\linewidth]{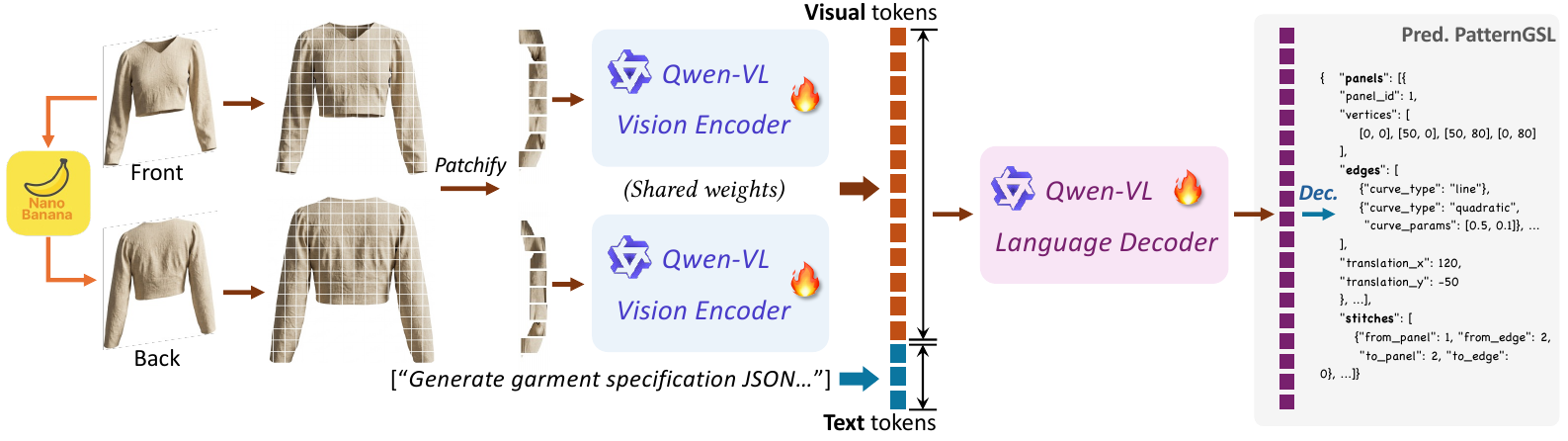}
    \caption{VLM architecture for image-to-pattern prediction. Given a front-view image, NanoBanana synthesizes the back view. Both views are patchified and processed by shared-weight Qwen-VL Vision Encoders to produce visual tokens. These are concatenated with text tokens from the instruction prompt and fed into Qwen-VL Language Decoder, which autoregressively generates PatternGSL tokens. A decoder (Dec.) converts the predicted tokens into structured JSON format. Fire icons indicate trainable components.}
    \label{fig:vlm}
\end{figure*}

\section{Image-to-Pattern Prediction via Vision-Language Model}
\label{sec:application}

{We use \methodname{} as the output space for single-image sewing-pattern prediction.}

\subsection{Problem Formulation}
\label{sec:problem_formulation}

{As shown in Fig.~\ref{fig:vlm}, given a front-view RGB image, NanoBanana synthesizes an auxiliary back view, yielding a paired input $\mathcal{I} \in \mathbb{R}^{2 \times 3 \times H_{\text{img}} \times W_{\text{img}}}$. We predict the complete GSL specification, including panel vertices, curve parameters, 3D transforms, stitches, and the scale token. The task is conditional sequence generation:}
\begin{align}
    P(J | \mathcal{I}) = \prod_{t=1}^{T} P(j_t | j_{<t}, \mathcal{I}; \theta),
\end{align}
where $J = \{j_1, \ldots, j_T\}$ represents the GSL tokens output and $\theta$ denotes network's parameters. This formulation naturally handles variable garment complexity through variable-length sequences while maintaining structured output constraints.

\subsection{Model Architecture}
\label{sec:architecture}

{We employ Qwen3-VL~\cite{Qwen3-VL}, consisting of a vision encoder and language decoder. The front and synthesized back views are patchified and encoded independently with shared weights, producing $\mathcal{T}_{\text{front}}$ and $\mathcal{T}_{\text{back}}$. They are fused only in the language decoder after concatenation with the instruction tokens. Thus the synthesized back view is an auxiliary cue for hidden regions, rather than a hard geometric constraint imposed before generation.}

\emph{Vision Encoder.}
{The front view and synthesized back view are partitioned into flattened image patches $\mathcal{I}_{p} \in \mathbb{R}^{2 \times n \times l}$, where $n$ is the number of patches per image and $l$ is the number of pixel values in each patch. A shared-weight ViT encoder maps these patches to visual tokens, $\mathcal{T}_{\text{front}}=\{\mathbf{f}^{\text{front}}_1,\ldots,\mathbf{f}^{\text{front}}_n\}$ and $\mathcal{T}_{\text{back}}=\{\mathbf{f}^{\text{back}}_1,\ldots,\mathbf{f}^{\text{back}}_n\}$, with each token summarizing local appearance and shape cues. Processing the two views independently avoids imposing an early geometric alignment; cross-view reasoning is left to the language decoder during structured generation.}

\emph{Language Decoder.}
{Given visual tokens and a prompt such as ``Generate garment specification JSON'', the decoder autoregressively generates GSL tokens until the terminal $\langle$end$\rangle$ token is produced. Variable-length generation naturally supports garments with different numbers of panels, edges, and stitches.}
\subsection{\datasetname{} Dataset Construction}
\label{sec:dataset}
{To our best knowledge, no existing dataset offers large-scale paired image-pattern annotations for learning-based garment reconstruction. We introduce \datasetname{}, organized for two-stage training (Fig.~\ref{fig:pipeline}, bottom-left): (1) 250K synthetic rendered images generated from GarmentCodeData~\cite{korosteleva2024garmentcodedata} specifications with diverse viewpoints and lighting; and (2) 50K photorealistic images produced by NanoBanana for domain adaptation.
NanoBanana is used as an image-to-image translation and view-synthesis module: it transfers synthetic renderings toward more realistic textures and lighting and synthesizes a back view from the available front-view image.
The REAL subset is not manually annotated from unconstrained photographs. Instead, garments with known sewing patterns are rendered and translated to more realistic appearance, so the ground-truth GSL annotations are inherited from the original structured specifications. Truly in-the-wild photographs are reserved for evaluation.}

\subsection{Training Strategy}
\label{sec:training}

Inspired by Visual Instruct Tuning~\cite{liu2023visual}, we reformulate garment parsing as a visual instruction-following task. The fire icons in Fig.~\ref{fig:pipeline} indicate trainable model components during our two-stage training process. We aim to train a large vision-language model to generate structured garment sequences conditioned on user instructions. 
The dataset is constructed using a single-turn dialogue template, where the user input concatenates visual tokens with a textual command: ``$\langle$image$\rangle$ $\cdots$ $\langle$/image$\rangle$ Generate the garment specification JSON according to the input images.'' The system assistant's response is filled with the ground-truth GSL sequence.

We fine-tune Qwen3-VL using a standard cross-entropy loss: $\mathcal{L} = -\frac{1}{T} \sum_{t=1}^{T} \log P(j_t^{\text{gt}} | j_{<t}^{\text{gt}}, \mathcal{I}; \theta)$, where $j_t^{\text{gt}}$ denote ground-truth tokens. This provides direct supervision on structured outputs. At inference time (Fig.~\ref{fig:pipeline}, bottom-right), the frozen model predicts PatternGSL from in-the-wild images, which are then decoded and simulated to produce 3D draped garments.

\section{Experiments}
\label{sec:experiments}
We validate our approach through: (1) \emph{quantitative comparison} on 2D pattern accuracy and 3D simulation quality; (2) \emph{ablation studies} on boundary sampling and VLM training; (3) \emph{pattern editing} via panel scaling, curve adjustment, and component removal; (4) \emph{qualitative comparison}; and (5) \emph{in-the-wild generalization}.

\subsection{Experimental Configuration}
\label{sec:setup}

\paragraph{Dataset.} We use \datasetname{} described in Sec.~\ref{sec:dataset}, with a train/validation split. The validation set includes garments with novel topologies unseen during training. {On the pre-tokenization PatternGSL JSON split, the dataset contains 257,956 garments, 2.84M panels, and 8.17M stitches; edges per panel average 6.91 and stitches per garment average 31.66, with long tails up to 40 edges and 106 stitches. Full topology statistics are provided in the supplementary material.}
\paragraph{Baselines.} We compare against: (1) \textbf{SewFormer}~\cite{liu2023sewformer}, transformer-based for sewing pattern prediction; (2) \textbf{LGM}~\cite{tang2024lgm}, large multi-view Gaussian model for 3D generation; (3) \textbf{GarmentX}~\cite{guo2025garmentx}, autoregressive method for garment generation; (4) \textbf{GarmentImage}~\cite{li2024garmentimage}, which uses raster representation with CNN prediction. {Dress-1-to-3~\cite{li2025dress1to3} targets 3D garment recovery rather than explicit 2D sewing-pattern prediction; we discuss this output-space difference in the supplementary material.}

\paragraph{Metrics and Simulation.} We evaluate both 2D pattern quality (2D Chamfer distance in mm, 2D IoU, stitch connection accuracy) and 3D simulation quality (draping success rate, 3D Chamfer distance after simulation). {All methods use one fixed cloth protocol with global settings, not tuned per sample, and no extra learned pose-estimation module. Key parameters are sim\_fps=60, sim\_substeps=10, and max\_sim\_steps=2400; the full protocol is in the supplementary material.}

\subsection{Quantitative Comparison}
\label{sec:exp_quantitative}

We evaluate on two complementary aspects: 2D pattern reconstruction accuracy and 3D simulation quality. Table~\ref{tab:main} compares pattern-level metrics against baselines on the validation set. 
Our approach achieves substantially lower 2D Chamfer distance (5.78~mm vs. 29.07~mm for the next best) and higher 2D IoU (86.34\% vs. 58.50\%), indicating  superior geometric fidelity. The 98.48\% stitch accuracy confirms that our explicit topology encoding preserves connectivity information that implicit representations cannot capture.

\begin{table}[tbp]
\centering
\footnotesize
\caption{Quantitative comparison on the validation set.}
\label{tab:main}
\begin{tabular}{@{}lccc@{}}
\toprule
Method & 2D Chamfer (mm)$\downarrow$ & 2D IoU (\%)$\uparrow$ & Stitch Acc (\%)$\uparrow$ \\
\midrule
SewFormer & 108.22 & 20.70 & 0.04 \\
GarmentX & 35.07 & 58.30 & 38.81 \\
GarmentImage & 29.07 & 58.50 & N/A \\
\textbf{Ours} & \textbf{5.78} & \textbf{86.34} & \textbf{98.48} \\
\bottomrule
\end{tabular}
\end{table}

{Beyond 2D accuracy, Table~\ref{tab:simulation} evaluates draping success rate and 3D Chamfer distance after physics-based simulation.} Our method achieves 99.2\% success rate compared to 4.7\% for GarmentImage, validating that sub-millimeter coordinate precision ($\epsilon < 0.04$~mm) and explicit topology encoding satisfy the simulator constraints in most cases. Note that LGM generates 3D meshes rather than sewing patterns, so draping success rate is not applicable; we report only its 3D Chamfer distance for geometric comparison. {The low 3D Chamfer distance (6.31~mm) further confirms that predicted patterns produce realistic draping behavior without optimization-based refinement or manual cleanup. Figure~\ref{fig:qualitative} shows generalization to novel topologies unseen during training, including garments with over 12 panels and complex asymmetric designs.}

\begin{table}[tbp]
\centering
\footnotesize
\caption{Simulation quality assessment.}
\label{tab:simulation}
\begin{tabular}{@{}lcc@{}}
\toprule
Method & Success Rate (\%)$\uparrow$ & 3D Chamfer (mm)$\downarrow$ \\
\midrule
SewFormer & 81.2 & 666.77 \\
LGM & N/A & 26.79 \\
GarmentX & 95.3 & 68.62 \\
GarmentImage & 4.7 & 139.49 \\
\textbf{Ours} & \textbf{99.2} & \textbf{6.31} \\
\bottomrule
\end{tabular}
\end{table}

{We further measure generation reliability before and after deterministic validation. The raw JSON validity rate is 100.0\%, the strict no-repair simulator-ready rate is 95.31\%, and the final simulator-ready rate after deterministic validation is 99.2\%. Strict no-repair failures are dominated by seam and topology inconsistencies and degenerate triangles rather than syntax errors, with the full breakdown in the supplementary material.}

{We also evaluate robustness to the auxiliary back-view cue. In a unified setting where the synthesized back view is unavailable or unreliable, the model still achieves 12.23~mm 2D Chamfer, 85.14\% stitch accuracy, and 97.78\% simulation success. In a stronger mismatch setting where the synthesized back view is inconsistent with the front view in sewing pattern, color, or style, performance degrades to 31.35~mm, 50.57\%, and 66.18\%, respectively, indicating graceful degradation under severe hidden-view ambiguity.}

\subsection{Ablation Studies}
\label{sec:ablation}

We conduct ablation studies to validate key design choices in both representation design and VLM training configuration.

\paragraph{Representation Design: Boundary Sampling.}
Table~\ref{tab:ablation_sample} ablates our boundary sampling strategy, providing explicit geometric supervision for curved edges. We vary both sample count $N_s \in \{4, 8, 16\}$ and sampling scope (curved edges only vs. all edges). 
Results show that $N_s=8$ achieves optimal balance between geometric precision and token efficiency: fewer samples under-constrain curve shapes, while more samples mainly increase sequence length without proportional accuracy gains. Notably, sampling only curved edges outperforms sampling all edges, validating our design intuition that straight edges do not require explicit  geometric supervision and extra samples on them contribute noise rather than useful signal.

\begin{table}[htbp]
    \centering
    \footnotesize
    \caption{Ablation on boundary sampling strategy.}
    \label{tab:ablation_sample}
    \begin{tabular}{@{}lcc@{}}
    \toprule
    Configuration & 2D Chamfer (mm)$\downarrow$ & 3D Chamfer (mm)$\downarrow$ \\
    \midrule
    \multicolumn{3}{l}{\textit{Sample count (curved edges only):}} \\
    $N_s=4$ & 8.42 & 9.15 \\
    \textbf{$N_s=8$ (Ours)} & \textbf{5.78} & \textbf{6.31} \\
    $N_s=16$ & 6.03 & 6.58 \\
    \midrule
    \multicolumn{3}{l}{\textit{Sampling scope ($N_s=8$):}} \\
    All edges & 7.21 & 7.89 \\
    \textbf{Curved edges only (Ours)} & \textbf{5.78} & \textbf{6.31} \\
    \bottomrule
    \end{tabular}
\end{table}

\paragraph{VLM Training Configuration.}
Table~\ref{tab:vlm_ablation} ablates model capacity and fine-tuning strategies. Full fine-tuning of the 8B model with an unfrozen vision encoder achieves best performance. 
LoRA-based fine-tuning underperforms full fine-tuning, suggesting that accurate GSL generation requires adapting both visual understanding and language generation capabilities---the structured output format demands tight coordination between perceiving garment geometry and generating valid JSON sequences.
Freezing the vision encoder also degrades performance, indicating that generic pre-trained visual features are suboptimal for extracting garment-specific geometric cues and benefit from task-specific adaptation.

\subsection{Pattern Editing}
\label{sec:pattern_editing}

{A key advantage of explicit parameterization is the ability to edit predicted patterns directly. Since GSL stores geometric parameters with semantic meaning, designers can modify the JSON specification and run the same deterministic decoder.}

\begin{figure*}[t]
    \centering
    \includegraphics[width=0.85\linewidth]{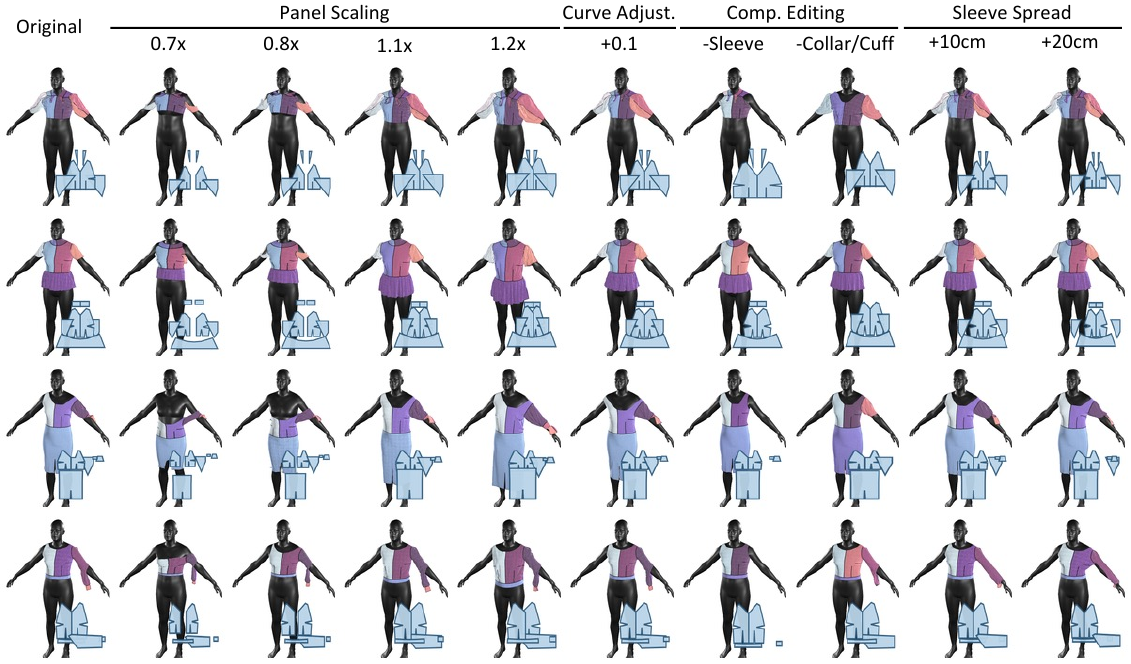}
    \caption{Pattern editing via direct \methodname  manipulation. Each row shows one garment undergoing systematic edits: \emph{Panel Scaling} (0.7$\times$--1.2$\times$) uniformly resizes all panels; \emph{Curve Adjustment} reshapes boundaries via B\'{e}zier curvature; \emph{Component Removal} eliminates semantic parts (sleeves or collars/cuffs); \emph{Sleeve Spread} (+10cm, +20cm) adjusts panel layout by spreading sleeves outward. Inset patterns visualize the 2D sewing layout. All 40 edited configurations produce valid, simulation-ready outputs with 100\% draping success, demonstrating that our explicit parameterization preserves geometric consistency under arbitrary modifications.}
    \label{fig:editing}
\end{figure*}

{We implement four common operations, illustrated in Fig.~\ref{fig:editing}: panel scaling, curve adjustment, component removal, and sleeve spread. Each operation modifies explicit GSL fields such as vertices, curve parameters, panel labels, or translations; mathematical details are provided in the supplementary material.}

{Figure~\ref{fig:editing} demonstrates these operations across four garment samples with diverse topologies. Panel scaling (0.7--1.2$\times$), curve adjustment, component removal, and sleeve spread (+10cm, +20cm) produce corresponding changes in both 2D layouts and 3D drapes. Across 40 edited patterns, all cases pass physics-based simulation, achieving 100\% draping success. Additional editing formulations are provided in the supplementary material.}

\begin{table}[tbp]
    \centering
    \footnotesize
    \caption{Ablation on VLM training configurations.}
    \label{tab:vlm_ablation}
    \begin{tabular}{@{}lccc@{}}
    \toprule
    Configuration & 2D IoU$\uparrow$ & 2D Chamfer$\downarrow$ & 3D Chamfer$\downarrow$ \\
    \midrule
    4B w/ LoRA & 71.46 & 26.82 & 72.35 \\
    4B w/ Freeze & 66.79 & 32.39 & 124.79 \\
    4B w/o Freeze & 84.16 & 13.57 & 36.64 \\
    \textbf{8B w/o Freeze} & \textbf{85.14} & \textbf{12.38} & \textbf{14.47} \\
    \bottomrule
    \end{tabular}
\end{table}

\subsection{Qualitative Results} 
\begin{figure*}[t]
    \centering
    \includegraphics[width=0.98\linewidth]{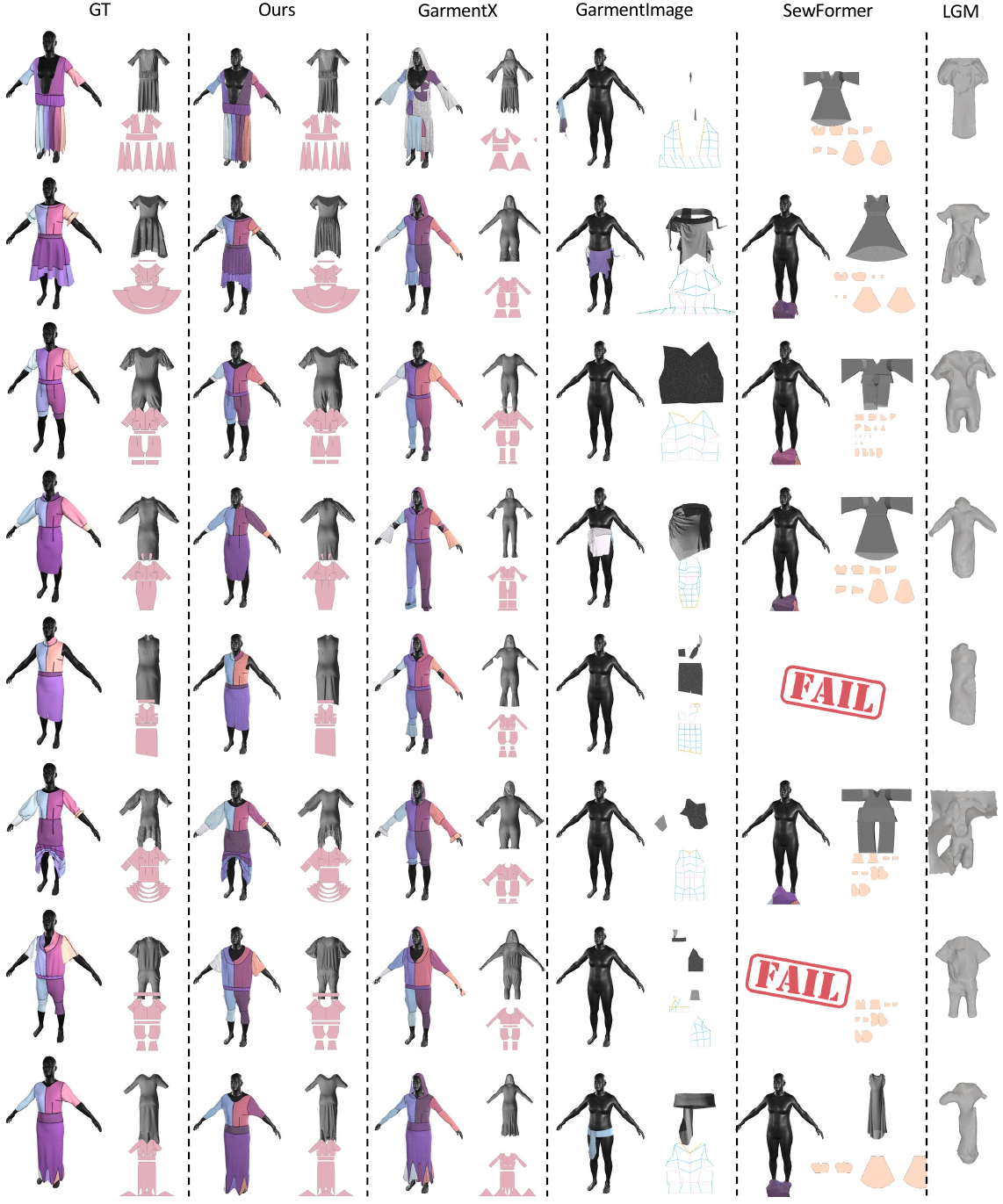}
    \caption{Qualitative comparison with baselines. For each sample, we show 3D draping results and 2D sewing patterns. Our method produces accurate patterns with realistic draping across diverse garment types. GarmentImage frequently fails to drape correctly due to imprecise geometry. SewFormer fails on complex topologies (rows 4 and 6). LGM generates 3D meshes directly without sewing patterns.}
    \label{fig:qualitative}
\end{figure*}

Figure~\ref{fig:qualitative} presents side-by-side comparisons across seven garment samples with diverse topologies. Our method consistently produces accurate sewing patterns that closely match ground truth geometry and drape realistically on the human body. GarmentX generates reasonable patterns but exhibits geometric distortions in complex regions such as sleeve-body connections. GarmentImage suffers from severe draping failures across most samples---the predicted patterns fail to properly cover the body due to imprecise geometry and invalid stitch topology, confirming the quantitative finding of only 4.7\% simulation success rate. SewFormer completely fails on two samples (rows 4 and 6), producing garbled patterns that cannot be simulated. LGM generates plausible 3D meshes but lacks sewing pattern output entirely, limiting its utility for downstream manufacturing applications. These results demonstrate that our explicit GSL encoding enables both accurate pattern reconstruction and reliable physics-based simulation.

\subsection{In-the-Wild Generalization}
\label{sec:in_the_wild}

\begin{figure*}[t]
    \centering
    \includegraphics[width=0.88\linewidth]{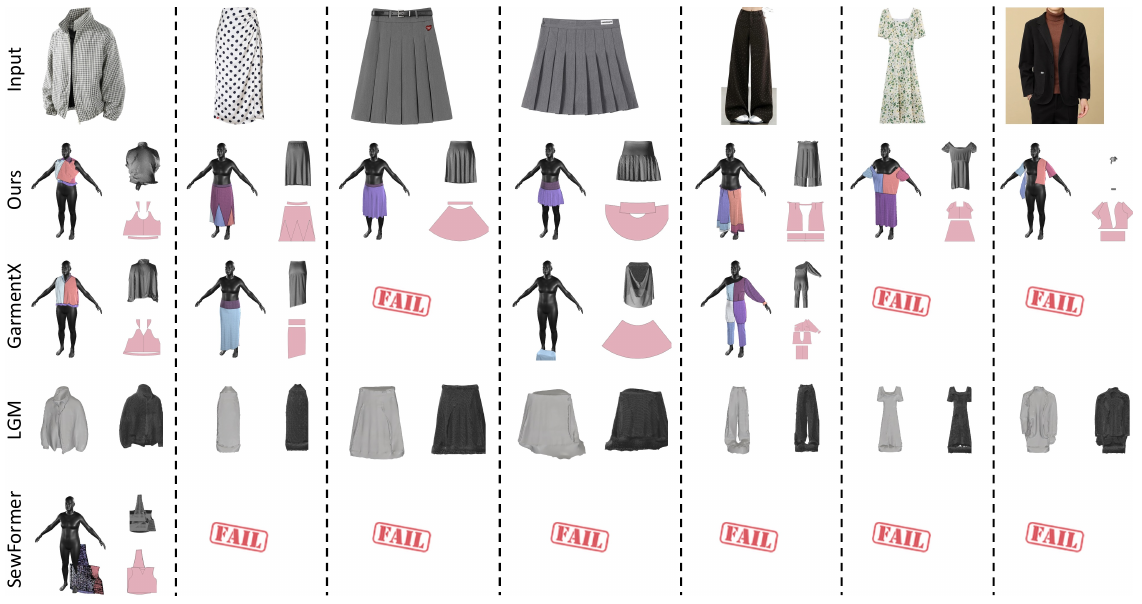}
    \caption{In-the-wild generalization across seven garment samples. Each column shows an input image (row 1) and results from four methods: Ours (row 2), GarmentX (row 3), LGM (row 4), and SewFormer (row 5). For pattern-based methods, we display both 3D draping and 2D sewing pattern; LGM outputs only 3D meshes without patterns. ``FAIL'' indicates simulation failure. Our method achieves 100\% success, while GarmentX fails on 3/7 and SewFormer fails on 6/7 samples. \textbf{More qualitative comparisons are in the supplementary materials}.}
    \label{fig:generalization}
\end{figure*}

To validate practical applicability beyond synthetic data, we evaluate on in-the-wild garment images collected from e-commerce websites and fashion photography. These images exhibit significant domain shift from training data, including complex backgrounds, varied lighting conditions, occlusions, and diverse garment styles unseen during training.
Figure~\ref{fig:generalization} compares our method against baselines on seven in-the-wild samples. Our model achieves 100\% simulation success, correctly inferring panel geometry and stitch topology despite challenging imaging conditions. In contrast, SewFormer fails on 6/7 samples due to garbled pattern outputs, while GarmentX fails on 3/7 samples, struggling with pants and complex dresses. LGM produces plausible 3D meshes but lacks sewing pattern output entirely. {Extended evaluation on 524 in-the-wild samples with comprehensive baseline comparisons is provided in the supplementary material.}


\section{Conclusion}
\label{sec:conclusion}

{We have presented \methodname{}, a structured representation for image-to-pattern reconstruction, along with \datasetname{}, a large-scale dataset of 300K paired samples for supervised VLM training. 
\methodname{} encodes geometric and topological sewing information in a compact format, achieving a 2D Chamfer distance of 5.78~mm and 99.2\% draping success.
The explicit parameterization further enables pattern-level editing applications -- including panel scaling, curve adjustment, and component editing -- while preserving compatibility with the deterministic decoding and simulation pipeline.}

\paragraph{Limitations and Future Work.}
{Our current evaluation covers an empirical topology range of 2--37 panels; higher panel counts and longer stitch sequences may require stronger long-context generation or hierarchical decoding. The synthesized back view is useful for hidden regions, and our robustness analysis shows graceful degradation when this cue is missing or inconsistent, but severe front-back ambiguity remains challenging. The current simulator also focuses on closed garments and static draping. Future work includes reducing seam and topology failures directly during generation, extending to open-boundary garments, and modeling dynamic cloth behavior.}


\clearpage

\bibliographystyle{ACM-Reference-Format}
\bibliography{main}


\end{document}